\documentclass[sigconf,nonacm]{acmart}

\setcopyright{none}
\renewcommand\footnotetextcopyrightpermission[1]{}
\usepackage{booktabs}
\usepackage{pgfplots}
\pgfplotsset{compat=1.18}
\begin{document}

\title{Certifying What Helps Customer-Return Timing: A Screen-and-Confirm Test for Conditioning Signals, and Why Decay Is Nearly Enough}

\author{Sang Su Lee}
\affiliation{%
  \institution{Thumbtack, Inc.}
  \city{San Francisco}
  \state{CA}
  \country{USA}
}
\email{psulee@thumbtack.com}

\author{Vineeth Loganathan}
\affiliation{%
  \institution{Thumbtack, Inc.}
  \city{San Francisco}
  \state{CA}
  \country{USA}
}
\email{vloganathan@thumbtack.com}

\author{Shishir Dash}
\affiliation{%
  \institution{Thumbtack, Inc.}
  \city{San Francisco}
  \state{CA}
  \country{USA}
}
\email{shishirdash@thumbtack.com}

\author{Vijay Raghavan}
\affiliation{%
  \institution{Thumbtack, Inc.}
  \city{San Francisco}
  \state{CA}
  \country{USA}
}
\email{vraghavan@thumbtack.com}

\renewcommand{\shortauthors}{Lee et al.}

\begin{abstract}
Practitioners enrich customer-return models with ever more signals (lifetime value, category, recency/frequency, calendar, geography), and the temporal-point-process (TPP) literature follows suit with covariate- and external-covariate-conditioned intensities. But \emph{does any of it improve the timing}, and how would you know? A null (``feature $X$ doesn't help'') is only meaningful if the model \emph{could} have found a signal. We make two contributions---a method and a measurement---to answer this credibly. \textbf{(i)} A \textbf{screen-and-confirm protocol} that \emph{certifies} whether a candidate signal improves a TPP's event-timing likelihood: a positive control plants a coupling of known strength and confirms the model recovers it, so a real-data null can be read as ``no signal'' rather than ``weak method.'' The control is validated for categorical and continuous encodings, and on a real clock-driven dataset (NYC taxi hour-of-day). \textbf{(ii)} A \textbf{model-free ceiling} quantifying how little of customer-return timing is \emph{point}-predictable at all ($\lesssim\!5\%$ of gap variance from any covariate; returns are near-memoryless). With these we certify a clean result on three public benchmarks (Amazon, Taobao, RetailRocket) and a real marketplace (Thumbtack): the inter-event clock---continuous-time \emph{decay}, long known to beat frozen-intensity models---is nearly \emph{sufficient}, and the conditioning the field keeps adding is redundant or harmful on top of it (statistically null on the public benchmarks, $\lesssim\!0.06$ NLL; null to mildly harmful on the marketplace). We do \emph{not} claim to discover that decay helps; our contribution is the tools that turn ``conditioning doesn't help'' into a checkable, certified statement---plus an honest-evaluation account of the read-out/leakage pitfalls we hit and retracted.
\end{abstract}


\maketitle

\renewcommand{\thefootnote}{}
\footnotetext{Accepted at the 5th Workshop on End-to-End Customer Journey Optimization (KDD 2026).}
\renewcommand{\thefootnote}{\arabic{footnote}}

\section{Introduction}
``When will this customer come back?'' underlies notification timing, CRM prioritization, and lifecycle modeling. Temporal point processes (TPPs) answer it by modeling the conditional intensity $\lambda^*(t\mid\mathcal{H}_t)$---the instantaneous event rate given the past. Neural TPPs (RMTPP~\cite{rmtpp} $\to$ NHP~\cite{nhp} $\to$ Transformer-Hawkes/SAHP~\cite{thp,sahp} $\to$ state-space models~\cite{s2p2}) have steadily improved likelihood fit. But a practitioner choosing a model for customer return faces a question the literature does not answer cleanly: \textbf{which components actually move return-timing, and how would you know?} Enriching event models with such signals is common practice: marked TPPs condition on event type/category by design~\cite{rmtpp,nhp}; covariate TPPs add feature vectors and even learn their importance~\cite{transfeat}; external-covariate TPPs inject seasonal/periodic drivers into the intensity~\cite{metp}; and recency/frequency (RFM) and value features power deep churn and lifetime-value models~\cite{ziln}. These works typically \emph{add} a signal and report a gain, in domains where it plausibly drives timing. We ask a different question for customer-\emph{return} timing: not whether such a covariate \emph{can} be added, but whether it adds anything once a strong inter-event backbone (decay) is already in place---and how to tell a true null from a method too weak to find the signal.

We answer empirically on three public benchmarks (Amazon, Taobao, RetailRocket) plus a real marketplace (Thumbtack), reporting temporal negative log-likelihood (NLL), inter-event RMSE, and MAE. Our contributions:
\begin{itemize}
  \item \textbf{C1. A screen-and-confirm certification (method).} A protocol that certifies whether a candidate signal improves a TPP's event-timing likelihood: a positive control plants a coupling of known strength and confirms recovery (categorical \emph{and} continuous encodings, plus a real clock-driven dataset, NYC taxi), so a real-data null reads as ``no signal,'' not ``weak method'' (\S\ref{sec:screen}).
  \item \textbf{C2. A model-free ceiling (measurement).} How little of customer-return timing is \emph{point}-predictable at all: $\lesssim\!5\%$ of gap variance from any covariate; returns are near-memoryless (\S\ref{sec:ceiling}).
  \item \textbf{C3. The certified result.} Using C1--C2, on three public benchmarks and a real marketplace: continuous-time \emph{decay} (the inter-event clock) is nearly \emph{sufficient}, and the conditioning practitioners keep adding---LTV, category, RFM, calendar, geography---is redundant or harmful on top of it (statistically null on the public benchmarks, $\lesssim\!0.06$ NLL; null-to-harmful, up to $+0.65$, on the marketplace), for a \emph{dual mechanism} reason (\S\ref{sec:decay},\,\S\ref{sec:cond},\,\S\ref{sec:mech}).
\end{itemize}
\textbf{The two contributions are one story.} Our contribution is the \emph{tools}---a screen-and-confirm certification (C1) and a model-free ceiling (C2); decay is a known mechanism, not our finding. The result (C3, decay nearly suffices) and the tools are inseparable: ``conditioning doesn't help'' is trustworthy \emph{only} because the positive control shows the null means ``no signal,'' not ``a method too weak to find it.'' The tool is what licenses the result. (Plain THP, our frozen-intensity reference, is an \emph{ablation}, not a deployment target.)

The transferable lesson: for return-\emph{timing}, the inter-event clock (decay) is the reliable lever, and \textbf{before believing that a feature helps, screen it against a positive control}---so that a null means ``no signal in the data,'' not ``a method too weak to find it.''

\section{Related Work}
\label{sec:related}
\textbf{Neural TPP methods.} RMTPP~\cite{rmtpp} $\to$ NHP~\cite{nhp} $\to$ attention models THP~\cite{thp} and SAHP~\cite{sahp} $\to$ AttNHP~\cite{attnhp}; intensity-free density models~\cite{intfree}; recent continuous-time state-space / latent-linear-Hawkes models (S2P2~\cite{s2p2}). EasyTPP~\cite{easytpp} is the standard benchmark/codebase; the field evaluates on NLL, RMSE/MAE, and mark accuracy~\cite{review}. We compare these backbones plus conditioning components on the customer-return task.

\textbf{Customer return / CLV.} Buy-till-you-die models (BG/NBD \cite{bgnbd}; Pareto/NBD \cite{paretonbd}) and deep CLV~\cite{ziln} model value/return but not the full event-time intensity. Notably, BTYD models \emph{assume} Poisson purchasing while alive---exponential, memoryless inter-purchase gaps per customer; our model-free ceiling (\S\ref{sec:ceiling}) independently confirms that forty-year-old assumption on four datasets, and the small residual previous-gap regularity ($r^2$ $1$--$6\%$) is the mild timing regularity the Pareto/GGG extension models~\cite{paretoggg}. Hazard/point-process treatments of user return include Kapoor et al.~\cite{kapoor} and Du et al.~\cite{du2015}; the closest neural treatment is the RNN survival model of Grob et al.~\cite{grob}, which predicts a \emph{single} next return time per user. We instead model the full intensity $\lambda^*$ and ask \emph{which component matters on what data}, varying the backbone (NHP, THP, S2P2) and conditioning rather than committing to one architecture.

\textbf{Conditioning TPPs on covariates.} Injecting side information into the intensity is well studied: covariate TPPs encode feature vectors and learn per-feature importance (TransFeat-TPP~\cite{transfeat}); external-covariate TPPs decompose seasonal/periodic drivers across temporal granularities (METP~\cite{metp}); and marked TPPs~\cite{rmtpp,thp} condition on event type by construction. These report accuracy gains in settings where the covariate drives event dynamics. We do not propose a new conditioning mechanism; we measure whether the standard ones help for customer-return \emph{timing} against a decay backbone, and add a positive-control screen so a null is interpretable.

\textbf{Exogenous and causal TPPs.} Estimating whether an \emph{external} signal (calendar, geography, or an intervention such as marketing) drives event timing is, in general, a causal question; counterfactual/causal TPPs~\cite{counterfactualtpp} formalize it but require interventional or carefully-controlled data. Our screen-and-confirm test (\S\ref{sec:screen}) is an \emph{observational} necessary condition---a feature that fails to improve conditional likelihood when correctly supplied cannot be a useful timing signal---and we position the causal treatment as future work.

\section{Models}
We hold the conditioning framework fixed and vary the \textbf{backbone} and \textbf{components}.

\textbf{Backbones.} (i) \emph{NHP}: continuous-time LSTM (CTLSTM); the hidden state decays between events, but the LSTM attenuates long-range history. (ii) \emph{THP}: Transformer with causal self-attention over events; captures long-range dependence. In our \emph{plain THP} ablation the intensity is fully \emph{frozen} between events: we omit the original THP's current-influence term $\alpha(t-t_j)/t_j$~\cite{thp}, so published THP is only \emph{state}-frozen (its intensity is linearly time-modulated between events) while our ablation removes between-event time-dependence entirely. (iii) \emph{S2P2}: a state-space / latent-linear-Hawkes layer whose intensity integral is computed in \emph{closed form} (no Monte-Carlo error).

\textbf{Components} (on the THP backbone). (a) \emph{Decay head} (``-D''): let the hidden state decay between events, $h(t)=h(t_i)\exp(-\delta\,\Delta t)$ with a small learned $\delta$; this relaxes THP's frozen-intensity assumption. (b) \emph{LTV} gate/shift and (c) \emph{category-aware decay}: condition the intensity on a per-customer value signal and event category. (d) \emph{RFM}: a causal recency/frequency/cadence feature---frequency $=\log(1+\text{position})$ and cadence $=\log\mathrm{1p}$ of the causal prefix-mean of inter-event times (using only events up to position $i$, to avoid leaking the target). (e) \emph{Exogenous}: per-event season and region embeddings. (f) \emph{Continuous covariate} (``-W''): a learned linear projection that injects a per-event continuous covariate vector into the event representation (we call it the \emph{weather projection} after its original use case); it carries the continuous positive control and the taxi hour-of-day covariate (\S\ref{sec:screen}). We make \emph{no} architectural-novelty claim; these are standard components whose \emph{data-dependent value} we measure.

\section{Experimental Setup}
\textbf{Datasets.} Three public benchmarks---Amazon (product reviews, essentially one event type), Taobao (e-commerce actions, 17 types), RetailRocket (very long web-browsing sequences)---and \textbf{Thumbtack}, a U.S.\ home-services marketplace where a customer sends a \emph{paid request} to professionals for a job (we keep customers with $\geq\!3$ such requests; a customer's per-event ``region'' is their Designated Market Area, DMA---one of $\sim$210 U.S.\ media-market regions). Public benchmarks use the EasyTPP-Gatech splits; Thumbtack is proprietary and reported in \emph{relative} terms.

\textbf{Metrics.} Our primary metric is the \textbf{temporal NLL} ($\downarrow$). For an event-time sequence $t_1<\dots<t_n$ with conditional intensity $\lambda^*(t)=\lambda(t\mid\mathcal H_t)$, the temporal log-likelihood is
\begin{equation}
\log L = \sum_{i=1}^{n}\log\lambda^*(t_i) \;-\; \int_{t_0}^{t_n}\lambda^*(u)\,du ,
\label{eq:tppll}
\end{equation}
and we report the \emph{per-event} temporal NLL $=-\tfrac1n\log L$. It isolates return-\emph{timing}, is computed by one consistent pipeline across all datasets and model families (the integral in closed form for S2P2-F, by Monte-Carlo quadrature otherwise), and excludes the mark term. (Cross-model NLL therefore mixes a closed-form compensator for S2P2-F with an $n_{\mathrm{mc}}\!=\!10$ Monte-Carlo estimate for the others; we treat the S2P2-F-vs-MC NLL gap cautiously and rest our claims on the \emph{same-estimator} decay contrast (THP vs.\ THP-D, an identical Monte-Carlo compensator).) We also report \textbf{inter-event RMSE/MAE} ($\downarrow$); these are comparable \emph{within a dataset only}, as the time unit differs across datasets. The same training protocol (Adam, warmup, early stopping on validation NLL) is used for all models, with mean$\pm$std over seeds $\{42,123,456\}$. Throughout, we read a conditioning delta as \emph{null} when $|\Delta| < 2\times$ its seed std and as real otherwise; the taxi recovery of \S\ref{sec:screen} ($\approx\!3\times$) clears this bar, and no decay-regime conditioning delta does.

\textbf{Backbone validation.} Before drawing component conclusions we confirm our backbones reproduce known behavior: the relative ranking of S2P2, NHP, THP, and RMTPP matches the S2P2 paper~\cite{s2p2}. (Three further reimplementations did not reproduce their published likelihoods in our pipeline; we omit them rather than report numbers we cannot trust, with details in App.~\ref{app:repro}.)

\section{Results}
We present the evidence bottom-up. The decay gain (\S\ref{sec:decay}) and backbone comparison (\S\ref{sec:backbone}) establish that the inter-event clock dominates; the conditioning redundancy (\S\ref{sec:cond}) and the \textbf{model-free ceiling} (\S\ref{sec:ceiling}, contribution~C2) measure how little signal remains; and the \textbf{screen-and-confirm protocol} (\S\ref{sec:screen}, contribution~C1) is the method that makes the negatives credible. The two framed contributions thus arrive last by design---the earlier subsections build to them.

\subsection{Decay is the dominant timing lever (a known mechanism, quantified here)}
\label{sec:decay}
Adding the decay head to a plain Transformer-Hawkes (THP $\to$ THP-D) yields large temporal-NLL gains on \emph{every} dataset (Table~\ref{tab:decay}): Amazon $0.32\!\to\!-2.55$ ($\Delta\,{-}2.9$), RetailRocket $0.52\!\to\!-3.53$ ($\Delta\,{-}4.0$), Taobao $-2.09\!\to\!-2.58$ ($\Delta\,{-}0.5$); all mean over 3 seeds, std $\le0.09$ (Table~\ref{tab:backbone}); and on Thumbtack a $\Delta$ of $-1.26$. This gain is \emph{distributional}---it sharpens the timing \emph{likelihood}, not point error: with the point-prediction leak removed (\S\ref{sec:limits}), THP-D's RMSE/MAE match plain THP on every dataset (\S\ref{sec:ceiling}). An earlier large MAE reduction we reported here was that read-out artifact and is retracted; the point-prediction view is the model-free ceiling of \S\ref{sec:ceiling}. The gain's \emph{size} varies (small on Taobao, large elsewhere), but its \emph{sign} is negative on every dataset---this is the most consistent result in the study. Mechanism: decay relaxes THP's frozen-between-events intensity so ``time since last event'' drives the rate---the classical mechanism of Hawkes processes and the neural Hawkes process~\cite{hawkes,nhp}. \emph{Is the gain just decay, or THP?} Continuous-time models with built-in decay (NHP, and the state-space S2P2) likewise sit far below plain THP; the lever is the decay mechanism, not the backbone it is attached to. (Plain THP is a fair \emph{ablation}---identical training and tuning, the between-event decay the single controlled variable---not a model one would deploy; modern attention TPPs are not frozen, and our plain THP omits the original's current-influence term (\S3), so the contrast quantifies the full between-event time-dependence channel rather than a deficit of published THP.) The same mechanism underlies S2P2's strong likelihood in \S\ref{sec:backbone}---exponential decay is the dominant mode of its state-space relaxation dynamics---so ``decay is the lever'' spans architectures, not merely the THP head.

\begin{table}[t]
\caption{Decay gain: $\Delta$ temporal NLL from adding the decay head (THP\,$\to$\,THP-D); negative $=$ better. Mean over 3 seeds; Thumbtack relative. Negative on every dataset---size varies (smallest on Taobao), sign does not.}
\label{tab:decay}
\centering
\begin{tabular}{lcccc}
\toprule
 & Amazon & Taobao & RR & Thumbtack \\
\midrule
$\Delta$ temporal NLL & $-2.9$ & $-0.5$ & $-4.0$ & $-1.26$ \\
\bottomrule
\end{tabular}
\end{table}

\subsection{No single best backbone---data decides on likelihood}
\label{sec:backbone}
This comparison is \emph{not} itself a contribution; it plays two supporting roles---confirming our reproductions track the literature (Backbone validation, above) and locating the decay mechanism. Table~\ref{tab:backbone} reports temporal NLL (mean$\pm$std over three seeds) for the models we could seed cleanly. The robust, \emph{same-estimator} comparison is the \textbf{decay contrast} (THP$\to$THP-D, both Monte-Carlo): decay lowers NLL by whole nats on every dataset (std $\le 0.09$), confirming \S\ref{sec:decay} with error bars. The closed-form state-space S2P2-F is competitive-to-best on the public sets ($-3.44$ Taobao, $-4.95$ RR), but its compensator is \emph{exact} while the THP models' is Monte-Carlo, so we parenthesize it as a reference and do \emph{not} rank it against the MC models (its Thumbtack cell is unavailable). The Monte-Carlo S2P2 and NHP variants are omitted; single runs do not change the picture. We therefore make \emph{no} universal-backbone-winner claim: decay is the \emph{first-order} lever (whole nats), residual cross-backbone differences are \emph{second-order} and data-dependent, and ``sufficient'' means decay reaches near the attainable ceiling---not that the backbone is irrelevant. This is consistent with the large-scale finding of Bosser and Ben Taieb~\cite{bosser} that time-NLL differences across neural TPP architectures are small; our multi-nat contrasts arise from the deliberately fully-frozen ablation (\S3), not from cross-architecture spread. We also make no claim against IntensityFree/FullyNN/AttNHP (dropped; competitive in their own papers). (Apparent point-timing RMSE differences are largely a \emph{read-out} artifact, not a backbone property; \S\ref{sec:limits}.)

\begin{table*}[t]
\caption{Temporal NLL ($\downarrow$), mean$\pm$std over 3 seeds. Public absolute; Thumbtack relative to THP-D ($\Delta$). The same-estimator \emph{decay contrast} (THP vs.\ THP-D, both Monte-Carlo) is the robust comparison. $^{\dagger}$S2P2-F uses an exact closed-form compensator (parenthesized; \emph{not} ranked against the MC models; Thumbtack cell unavailable). Monte-Carlo S2P2 and NHP are omitted.}
\label{tab:backbone}
\begin{tabular}{l rrrr}
\toprule
Model & Amazon & Taobao & RR & tt1$^{\Delta}$ \\
\midrule
THP                       & $0.32\pm0.00$  & $-2.09\pm0.00$ & $0.52\pm0.09$  & $+1.26\pm0.01$ \\
THP-D (THP$+$decay)       & $-2.55\pm0.01$ & $-2.58\pm0.00$ & $-3.53\pm0.07$ & $0.00$ \\
S2P2-Base-F (cf.)$^{\dagger}$ & $(-0.57\pm0.00)$ & $(-3.44\pm0.00)$ & $(-4.95\pm0.00)$ & --- \\
\bottomrule
\end{tabular}
\end{table*}

\noindent\emph{Pre-specified sanity filter.} We apply one fixed exclusion rule, decided from data/model properties rather than the ranking we wished to see: \textbf{implausible-given-inputs} variants are dropped---e.g.\ browse-augmented models reporting NLL $-8$ to $-16$ \emph{including on Amazon, which contains no browse data}, so the ``gain'' cannot come from the claimed signal.

\subsection{Conditioning is redundant under decay}
\label{sec:cond}
Why test this at all? Because adding these signals is exactly what practitioners and the recent TPP literature do (\S\ref{sec:related} covers covariate-, marked-, and external-covariate-conditioned models); the screen and ceiling let us \emph{certify} whether it actually pays for return-timing. We add each such signal on top of the backbone (Table~\ref{tab:cond}). A \emph{weak} (no-decay) baseline does benefit from category (CatOnly $\Delta\,{-}0.73$). But \textbf{once the decay head is present, every conditioning signal is null or harmful}: category worsens it (CatOnly-D $+0.47$; ExogCat-D $+0.65$, the worst), exogenous calendar/geography is null (ExogTHP-D $+0.012$), RFM collapses to $\approx\!0$ once its target leak is removed (\S\ref{sec:limits}), and proxy-LTV does not improve likelihood (LTV-CTPP-D $+0.47$). \textbf{The same holds on the public benchmarks} (Table~\ref{tab:condpub}): under decay every signal moves temporal NLL by $\lesssim\!0.06$, while the weak (no-decay) baseline still gains from category. We distinguish two regimes the table conflates: on public data the decay-regime effect is statistically \emph{null} (redundancy---the signal carries no information the clock lacks); on Thumbtack adding category under decay is mildly \emph{harmful} ($+0.47$), which we read as extra capacity overfitting in the absence of signal rather than as negative information (the harmful deltas' large seed variance, $\pm0.11$--$0.17$ in Table~\ref{tab:cond}, supports the instability reading). We therefore treat LTV/category/RFM/exogenous here as \emph{measured components, not contributions}: under the signals available offline they are redundant with decay. (Whether \emph{real} customer-revenue LTV helps remains open---the offline proxy ceiling is low by construction.) To be explicit about scope: this is a claim about the per-customer \emph{timing likelihood conditional on the event history}---not about targeting, churn/propensity, mark/volume, or value models, where these same features remain informative (\S\ref{sec:screen}).

\begin{table}[t]
\caption{Conditioning $\Delta$ temporal NLL relative to the same backbone ($\downarrow$ better; $+$ is worse), Thumbtack (relative); mean over 3 seeds, $\pm$ the seed std of the $\Delta$. The harmful decay-regime deltas carry large seed variance, consistent with the instability/overfitting reading (\S\ref{sec:cond}); ExogTHP-D is within seed noise. Conditioning helps only the weak baseline; under decay it is null or harmful.}
\label{tab:cond}
\begin{tabular}{l l r}
\toprule
Variant & Signal added & $\Delta$ temporal NLL \\
\midrule
\multicolumn{3}{l}{\emph{over THP (no decay)}} \\
\quad CatOnly      & category            & $\mathbf{-0.73}\pm0.01$ \\
\quad ExogTHP      & season $+$ region   & $+0.007\pm0.005$ \\
\midrule
\multicolumn{3}{l}{\emph{over THP-D (with decay)}} \\
\quad CatOnly-D    & category            & $+0.47\pm0.11$ \\
\quad RFM-D        & recency/freq/cadence& $-0.01\pm0.01$ \\
\quad LTV-CTPP-D   & proxy LTV           & $+0.47\pm0.10$ \\
\quad ExogTHP-D    & season $+$ region   & $+0.012\pm0.016$ \\
\quad ExogCat-D    & season$+$region$+$cat & $+0.65\pm0.17$ \\
\bottomrule
\end{tabular}
\end{table}

\begin{table}[t]
\caption{Conditioning $\Delta$ temporal NLL on the \emph{public} benchmarks, confirming the Thumbtack pattern: under decay every signal is $\lesssim\!0.06$; the weak (no-decay) baseline still gains from category. Mean$\pm$seed-std of $\Delta$ over 3 seeds; every decay-regime delta is within $\sim\!1\sigma$ of zero (null by the \S4 rule).}
\label{tab:condpub}
\centering
\small
\setlength{\tabcolsep}{3.5pt}
\begin{tabular}{l ccc}
\toprule
$\Delta$ NLL vs.\ same backbone & Amazon & Taobao & RR \\
\midrule
CatOnly-D ($+$decay) & $+0.03\pm0.04$ & $+0.00\pm0.00$ & $-0.06\pm0.05$ \\
RFM-D ($+$decay)     & $+0.00\pm0.08$ & $-0.01\pm0.00$ & $-0.01\pm0.03$ \\
CatOnly (no decay)   & $-0.32\pm0.00$ & $-0.02\pm0.00$ & $-1.70\pm0.04$ \\
\bottomrule
\end{tabular}
\end{table}

\subsection{A model-free ceiling: returns are near-memoryless}
\label{sec:ceiling}
Why does conditioning fail? Because, model-free, there is almost nothing to condition \emph{on}. By a \emph{model-free ceiling} we mean an upper bound on predictability computed \emph{without} fitting any TPP---the fraction of gap variance a covariate explains in a simple regression---so it bounds the \emph{mean-shift (point-prediction)} signal available to any model. By the reconciliation argument below it does \emph{not} bound distributional (likelihood) gains; the NLL-space redundancy claim rests on the measured deltas (Tables~\ref{tab:cond},~\ref{tab:condpub}) and the screen (\S\ref{sec:screen}), with the ceiling explaining why \emph{point-prediction} gains, specifically, are unavailable. Table~\ref{tab:ceiling} shows that no available covariate explains more than a single-digit percentage of gap variance on \emph{any} of the four datasets (we report the fraction of \emph{log}-gap variance explained: $r^2$ for the continuous previous gap, one-way $\eta^2$ for categoricals): on Thumbtack the previous gap is the best single predictor at $\approx\!2.2\%$ ($r=0.147$), category $\approx\!1.7\%$, and season and DMA region explain effectively nothing ($\eta^2\approx 5\text{--}8\times10^{-4}$); on the public benchmarks the pattern repeats (previous gap $1.4$--$6.3\%$, event-type/mark $\eta^2\le1.1\%$). Two regularities: the ceiling is single-digit everywhere, and the best single covariate is \emph{always the previous gap}---i.e.\ the inter-event clock itself, sequence-derivable and already captured by decay. Nor is the ceiling low by construction: on a genuinely clock-driven dataset (NYC taxi, \S\ref{sec:screen}), the same model-free check on the exogenous hour-of-day covariate yields $\mathrm{corr}=0.15$ ($r^2\approx2\%$)---an order of magnitude above Thumbtack's season $\eta^2$---and the conditioning screen fires there; on customer-return data the seasonal signal simply is not present. Moreover, behavioral surge (a burst of sessions or a marketing touch) \emph{appears to carry near-zero lead-time}---it coincides with the return rather than preceding it by days---so it could not be used to predict a return ahead of time (an informal observation, not a measured result; \S\ref{sec:limits}). Return timing is dominated by an exogenous, near-memoryless latent need; the inter-event clock captured by decay is most of the recoverable signal. Precisely, the process is near-\emph{renewal}: the hazard depends strongly on time since the last event (that is what decay captures) but on little else---``near-memoryless'' throughout means memoryless \emph{beyond the current gap's clock}, not a constant hazard.

\begin{table}[t]
\caption{Model-free fraction of log-gap variance explained by each covariate ($r^2$ for the continuous previous gap; one-way $\eta^2$ for categoricals). Single-digit everywhere, and the best covariate is always the \emph{previous gap}---the clock that decay captures. Public columns: pooled splits, category $=$ event type (marks); season/region covariates are unavailable on the public benchmarks.}
\label{tab:ceiling}
\centering
\small
\begin{tabular}{l cccc}
\toprule
Covariate & Amazon & Taobao & RR & Thumbtack \\
\midrule
Previous gap ($r^2$)  & $1.4\%$ & $3.4\%$ & $6.3\%$ & $\approx 2.2\%$ \\
Category/mark ($\eta^2$) & $1.1\%$ & $0.4\%$ & $0.3\%$ & $\approx 1.7\%$ \\
Season (month, $\eta^2$) & --- & --- & --- & $\approx 5\times10^{-4}$ \\
Region (DMA, $\eta^2$)   & --- & --- & --- & $\approx 8\times10^{-4}$ \\
\bottomrule
\end{tabular}
\end{table}

\textbf{Reconciling with the decay gain (\S\ref{sec:decay}).} A natural objection: if returns are near-memoryless, how can decay lower NLL by $2$--$3$ nats? Because the two measure different things. The ceiling is \emph{point-prediction} variance---how much a covariate moves the conditional \emph{mean} gap. The decay gain is a \emph{distributional} (likelihood) improvement: even when the mean is nearly unpredictable, modeling the \emph{shape} of the intensity (the post-event refractory dip and its drift back, and the compensator $\int\lambda^*$) calibrates the timing density far better than a flat-rate baseline. Decay does not predict \emph{which} gap better (low $r^2$); it places probability mass over \emph{when} far better (large $\Delta$NLL). So the findings are consistent: little is point-predictable, yet correct temporal \emph{calibration} is worth a lot---and it comes from the inter-event clock, not from external covariates. To make this concrete: on a \emph{perfectly} memoryless process (i.i.d.\ $\mathrm{Exp}(\lambda)$ gaps, where \emph{no} covariate has any predictive power), a model whose rate is off by a constant factor $c$ pays $c-1-\ln c$ excess nats per event---$1.6$ at $c\!=\!4$, $6.7$ at $c\!=\!10$. Multi-nat NLL gains are thus attainable at \emph{zero} point-predictability: getting the rate \emph{scale/shape} right (what decay does, versus THP's frozen between-event rate) is worth nats even when \emph{which} gap is unpredictable. Empirically this is exactly what we observe: once the point-prediction leak is removed (\S\ref{sec:limits}), THP, THP-D, and LTV-CTHP-D sit at the \emph{same} inter-event RMSE (Amazon $\approx\!0.30$, Taobao $\approx\!0.14$, RetailRocket $\approx\!9.5$), indistinguishable from predicting the global-mean gap, and a plain RFM$\to$gradient-boosting regressor matches the neural TPPs on this point-prediction. Nothing beats the constant on \emph{which} gap, while decay still wins whole nats on \emph{when}---the point-prediction ceiling and the distributional decay gain are two faces of the same near-memoryless process.

\subsection{Screening exogenous features: synthetic recover, real reject}
\label{sec:screen}
\textbf{The protocol.} \textbf{(1)} Fix the candidate feature's \emph{encoding} and the prediction target (here, inter-event timing). \textbf{(2)} On synthetic data, plant a feature$\to$target coupling of tunable strength $\beta$ and confirm the conditioned model recovers it \emph{monotonically} in $\beta$---a positive control that calibrates sensitivity. \textbf{(3)} Run the \emph{identical} pipeline on the real feature. \textbf{(4)} Interpret \emph{within the control's scope}: given a passing control, a flat real result certifies ``no signal in that encoding,'' not a weak method. The pattern ports standard positive/negative-control practice---placebo and refutation tests in econometrics, sanity checks in ML~\cite{adebayo}---to TPP conditioning.

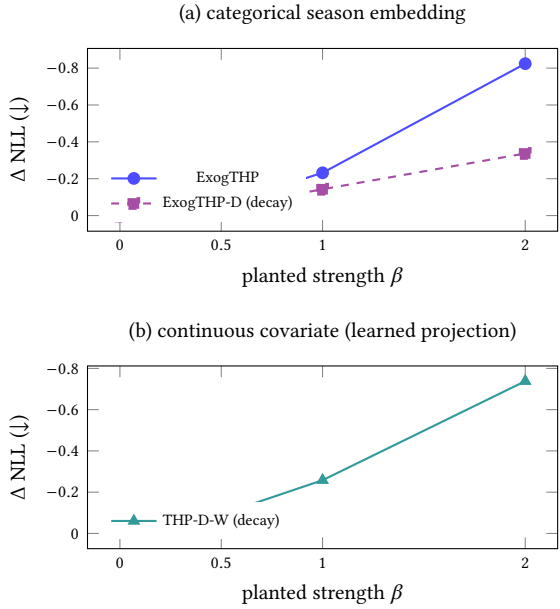
\begin{figure}[t]
\centering
\begin{tikzpicture}
\begin{axis}[
  width=0.92\columnwidth, height=4.0cm,
  title={(a) categorical season embedding},
  xlabel={planted strength $\beta$}, ylabel={$\Delta$ NLL ($\downarrow$)},
  y dir=reverse, xtick={0,0.5,1,2}, enlarge x limits=0.08,
  legend style={font=\scriptsize, at={(0.03,0.05)}, anchor=south west, draw=none},
  tick label style={font=\scriptsize}, label style={font=\small}, title style={font=\small},
  mark size=2pt,
]
\addplot[blue!70, thick, mark=*] coordinates {(0,0.001)(0.5,-0.059)(1,-0.232)(2,-0.824)};
\addplot[violet!70, thick, dashed, mark=square*] coordinates {(0,0.002)(0.5,-0.042)(1,-0.143)(2,-0.337)};
\legend{ExogTHP, ExogTHP-D (decay)}
\end{axis}
\end{tikzpicture}

\vspace{6pt}

\begin{tikzpicture}
\begin{axis}[
  width=0.92\columnwidth, height=4.0cm,
  title={(b) continuous covariate (learned projection)},
  xlabel={planted strength $\beta$}, ylabel={$\Delta$ NLL ($\downarrow$)},
  y dir=reverse, xtick={0,0.5,1,2}, enlarge x limits=0.08,
  legend style={font=\scriptsize, at={(0.03,0.05)}, anchor=south west, draw=none},
  tick label style={font=\scriptsize}, label style={font=\small}, title style={font=\small},
  mark size=2pt,
]
\addplot[teal!80, thick, mark=triangle*] coordinates {(0,0.0001)(0.5,-0.092)(1,-0.258)(2,-0.738)};
\legend{THP-D-W (decay)}
\end{axis}
\end{tikzpicture}
\caption{\textbf{Positive control (two encodings).} A planted exogenous$\to$gap coupling of strength $\beta$ is recovered \emph{monotonically} in both: \textbf{(a)} a \emph{categorical} season embedding (no-decay ExogTHP and decay ExogTHP-D), and \textbf{(b)} a \emph{continuous} covariate via a learned linear projection (decay THP-D-W). $\beta=0$ is correctly null in both, so the screen is not categorical-specific. This is what a \emph{real} signal would look like---real calendar/geography (Table~\ref{tab:cond}) does not.}
\label{fig:beta}
\end{figure}

A null is only informative if the method \emph{could} have found a signal. We therefore validate the exogenous pipeline with a \textbf{positive control}: we generate synthetic sequences in which an exogenous field (season) is coupled to the next gap at a tunable strength $\beta$ (a causal coupling that a decay-only model cannot recover, since the field is i.i.d.\ across events), and check that a season/region-conditioned model recovers it. It does, \emph{monotonically} in $\beta$ (Fig.~\ref{fig:beta}): ExogTHP improves by $-0.059$, $-0.232$, $-0.824$ at $\beta=0.5,1,2$ (and $+0.001$ at $\beta=0$, correctly null); the decay variant ExogTHP-D recovers as well ($-0.042,-0.143,-0.337$). The protocol is \emph{not} specific to categorical features: replacing the categorical season/region embedding with a learned linear projection of a per-event \emph{continuous} covariate $z$, and planting a continuous coupling $\text{gap}\sim\mathrm{Exp}(e^{\beta z})$ with $z\sim\mathcal N(0,1)$, yields the same monotonic recovery under decay---$\Delta$ time-NLL $+0.000,-0.092,-0.258,-0.738$ at $\beta=0,0.5,1,2$. Both instantiations are correctly null at $\beta=0$. The same pipeline, applied to real calendar/geography (Thumbtack), yields no improvement (Table~\ref{tab:cond}).

\textbf{The real null is genuine, not a plumbing artifact.} We verified that the exogenous signal reaches the model with rich variation: Thumbtack carries all 12 calendar months and 207 DMA regions ($99.8\%$ non-zero), well-distributed, with loader keys matching the data. Because the positive control confirms ``signal present $\Rightarrow$ recovered,'' the real-data null means ``nothing to recover''---calendar and geography do not move the per-customer conditional gap. This screen-and-confirm pattern is the recommended way to read a conditioning null.

\textbf{Scope of the claim.} The screen adjudicates a \emph{specific} pair: the candidate feature \emph{as encoded} and the chosen target (here, inter-event timing). A null therefore means ``no signal in that encoding for that target,'' bounded by what the positive control shows recoverable---\emph{not} ``no exogenous signal of any kind.'' Likewise, of the signals in C3, calendar/geography passed through this planted-coupling screen; the LTV/category/RFM nulls are \emph{measured} under the same pipeline and bounded by the same ceiling but were not separately screened---per-pathway positive controls are future work. This boundary is consistent with the broader literature: on canonical weather-sensitive data the exogenous covariate drives \emph{volume and marks} rather than per-event timing. As a model-free check, Beijing PM2.5 alert \emph{onsets} are near-uniform across calendar months and their inter-onset gaps are essentially season-invariant, and on the Walmart weather dataset temperature is uncorrelated with the inter-sale gap ($r\!=\!0.005$); a \emph{timing} screen is thus correctly null there, even though a mark/volume model benefits from the same covariate~\cite{transfeat,metp}. Extending the screen to continuous \emph{and} lagged encodings (each with its own positive control) is a direct generalization.

\textbf{A real positive control.} The screen also fires on a \emph{real} exogenous signal where one genuinely exists. On NYC green-taxi pickups (Jan 2019), inter-pickup gaps are modulated by \emph{hour-of-day} (rate high at rush hour, low overnight; model-free $\mathrm{corr}(\sin\text{hour},\log\text{gap})=0.15$). Encoding hour as a continuous cyclic covariate, the screen lowers temporal NLL by $\Delta\!=\!-0.025$ ($1.756\!\to\!1.731$; $n\!=\!3$, well outside seed noise $\pm0.008$; replicated on a second month, Feb 2019, $\Delta\!=\!-0.021$; hour is genuinely exogenous---each sequence starts at its zone-day's first pickup, so cumulative time does not encode the clock hour)---a \emph{modest} recovery that tracks the \emph{modest} real signal, in contrast to the strong synthetic $\beta$ and the flat ($\approx\!0$) real calendar/geography. This closes the loop: synthetic controls calibrate sensitivity, real negatives (PM2.5, Walmart) and a real positive (taxi) confirm the screen discriminates on real data---so the customer-return null is a property of that data, not a weakness of the method. We still describe screen-and-confirm as a \emph{validation discipline} (establish recoverability, then trust the null) rather than a turnkey discovery method.

\section{Mechanism: why decay absorbs the rest}
\label{sec:mech}
The redundancy has a \emph{dual mechanism}. \textbf{(A) Sequence-derivable signals are already captured.} RFM and category are deterministic functions of the observed event history, which the decay backbone already encodes; supplying them explicitly adds no information (RFM-D $\approx\!0$, CatOnly-D worse once decay is in). \textbf{(B) Exogenous signals do not move the conditional gap.} Season and region shift \emph{demand composition} (who buys what, where) but not the per-customer \emph{timing} of the next return, which is driven by an exogenous latent need realized same-day (\S\ref{sec:ceiling}). Timing $\neq$ demand-composition. Together these explain why a single inter-event-clock mechanism (decay) is sufficient and the usual enrichments are redundant.

\textbf{What this buys the practitioner.} For the notification-timing and CRM use cases of \S1: a decay-calibrated intensity supports return-\emph{window} estimation (when the hazard recovers from the post-event refractory dip) and send suppression during the dip; it does \emph{not} support point-timed sends (point prediction sits at the global-mean ceiling, \S\ref{sec:ceiling}), and the effect of \emph{timing an intervention} is a causal question requiring interventional data (\S\ref{sec:related}, future work).

\section{Limitations and honest evaluation}
\label{sec:limits}
We report results we initially got wrong and corrected, in the spirit of honest evaluation. An NHP target-indexing bug once produced a fake $7\times$ MAE improvement; an RFM cadence feature once leaked the target through a full-sequence normalizer; an S2P2 RMSE was read from an untrained auxiliary head; and the decay head's point-prediction read the hidden state \emph{after} decaying it by the actual next inter-event interval, leaking the very gap it predicts (a counterfactual that alters only the held-out gap moves the prediction in lock-step, $\mathrm{corr}\!\approx\!1$)---which had inflated the decay models' RMSE/MAE and the apparent point-timing decay gains. Each was caught by sanity-checking predictions against targets; we fixed the decay read-out to use the pre-decay (history-only) state, added a counterfactual regression test, and report only the corrected, leak-free numbers, which show \emph{decay-flat} RMSE/MAE. Temporal NLL---a density evaluated at the observed event times and treated identically across all models---is unaffected by these point-prediction read-out bugs. Separately, in earlier ranking experiments a TPP's apparent ranking ability was dominated by the score \emph{read-out} (intensity integral vs.\ a learned time head) and the held-out \emph{anchor} rather than the architecture; an apparent ``likelihood-vs-ranking decoupling'' did not survive re-validation, so we make \emph{no} ranking claim. Scope limits: one real dataset, reported only in relative terms; the customer filter ($\geq\!3$ paid requests) conditions on repeat return, so seasonality acting on acquisition or the \emph{first} return is excluded by construction; proxy-LTV understates real-LTV conditioning (open); and our S2P2 point-timing is read from an auxiliary head, so we do not treat its RMSE as representative of the method. Three claims rest on argument more than on a dedicated experiment, which we flag: the model-free \emph{season/region} ceilings (\S\ref{sec:ceiling}) are measured on Thumbtack only (the public benchmarks carry no calendar/geography covariates), and platforms with strong external seasonality may sit under a higher ceiling, so the one-day ceiling-plus-screen check should be re-run per dataset instead of assuming our $\lesssim\!5\%$ figure; the ``decay already encodes RFM/category'' mechanism (\S\ref{sec:mech}) is \emph{inferred}, not probed (no representation-probing experiment); and the ``surge has zero lead-time'' observation is from exploratory analysis without a dedicated figure.

\section{Conclusion}
Two tools, a model-free ceiling and a screen-and-confirm certification, let us verify a clean result for customer-return \emph{timing}: continuous-time decay (a known mechanism) is nearly \emph{sufficient}, and once it is present, the conditioning practitioners add is redundant or harmful. A model-free analysis shows why: returns are near-memoryless, with $\lesssim\!5\%$ of gap variance explainable (and, informally, near-zero surge lead-time). Because a conditioning null is only meaningful against a positive control, we recommend a \textbf{screen-and-confirm} protocol: confirm the pipeline recovers a planted synthetic signal, then trust the real-data null. Future work: a causal/counterfactual TPP treatment of \emph{interventions} (e.g.\ marketing), which needs interventional or randomized-holdout data; representation-probing (or mutual-information) experiments on the decay backbone's hidden state, to turn the \emph{inferred} ``decay already encodes RFM/category'' mechanism of \S\ref{sec:mech} into a measured one; and generalizing screen-and-confirm into automatic exogenous-feature discovery.

\appendix
\section{Reproducibility details}
\label{app:repro}
\textbf{Splits.} Public benchmarks use the EasyTPP-Gatech \cite{easytpp} train\slash dev\slash test splits; Thumbtack uses a chronological per-customer split, reported only in relative terms.
\textbf{Optimization (all models).} Adam, learning rate $10^{-3}$, linear warmup over 5 epochs then \texttt{Reduce\-LR\-On\-Plateau} (factor $0.5$, patience $5$, min lr $10^{-5}$), gradient clipping at $1.0$, up to $50$ epochs with early stopping (patience $10$) on validation temporal NLL. Hidden size $d_{\text{model}}=64$; batch size $64$ (RetailRocket $16$, sequences truncated to $512$). $\pm$ is mean/std over seeds $\{42,123,456\}$.
\textbf{Backbones.} THP: causal self-attention, intensity constant between events (the original's current-influence term $\alpha(t-t_j)/t_j$ is omitted; \S3); ``-D'' adds the per-dimension hidden-state decay head (small learned $\delta$). NHP: continuous-time LSTM. \textbf{S2P2 is our reimplementation} of a continuous-time state-space / latent-linear-Hawkes layer ($d_{\text{state}}=64$, $2$ layers); \emph{-F} computes $\int\lambda^*$ in closed form ($n_{\text{mc}}=0$), the non-F variant uses $n_{\text{mc}}=10$. We do not use the original authors' code.
\textbf{Dropped baselines.} IntensityFree~\cite{intfree}, FullyNN~\cite{fullynn}, and AttNHP~\cite{attnhp} reimplementations did not reproduce published likelihoods in our pipeline (e.g.\ Taobao IntensityFree LL $-1.43$ vs.\ $+1.318$ reported), so we exclude their numbers and cite the originals, which report competitive results.
\textbf{Synthetic positive control.} Per-event season $\sim\mathrm{Unif}\{1..12\}$ i.i.d.; $\text{gap}_{j+1}\sim\mathrm{Exp}(r_0\exp(\beta\sin(2\pi\,\text{season}_j/12)))$ using the \emph{source} event's season (causal); region is an i.i.d.\ noise control. A pre-training validation confirms the planted correlation grows with $\beta$. The continuous variant draws $z\sim\mathcal N(0,1)$ per event and sets $\text{gap}_{j+1}\sim\mathrm{Exp}(e^{\beta z_j})$, fed through the weather projection.
\textbf{Model-free ceiling (public columns of Table~\ref{tab:ceiling}).} Computed over pooled train\slash dev\slash test splits, on $\log$ of positive inter-event gaps: $r^2$ from the Pearson correlation of adjacent within-sequence gap pairs $(\log g_j,\log g_{j+1})$; $\eta^2$ from a one-way decomposition of $\log g_{j+1}$ grouped by the source event's type (\texttt{ceiling\_public.py}). Applied to the Thumbtack event data, the same script yields $r^2\!\approx\!2.0\%$ and category $\eta^2\!\approx\!1.1\%$, consistent in magnitude with the cohort-derived Thumbtack column of Table~\ref{tab:ceiling}.
\textbf{Real positive control (taxi).} NYC green-taxi pickups, Jan 2019; sequences are per (pickup-zone, day) with $\geq\!20$ pickups (capped at 200, $n\!=\!4000$); the per-event covariate is hour-of-day encoded as $[\sin(2\pi h/24),\cos(2\pi h/24),0]$ through the same weather projection. Events are single-type (timing only).


\end{document}